\documentclass[letterpaper, 10 pt, conference]{ieeeconf}  

\IEEEoverridecommandlockouts                              

\usepackage{graphics} 
\usepackage{amsmath} 
\usepackage{amssymb}  
\usepackage{graphicx}
\usepackage{multirow}
\usepackage{booktabs}
\RequirePackage{xcolor}
\title{\LARGE \bf
Physics-Aware Diffusion for LiDAR Point Cloud Densification
}

\author{Zeping Zhang$^{1}$ and Robert Laganière$^{2}$
\thanks{$^{1,2}$Zeping Zhang and Robert Laganière are with the VIVA Research Lab, Department of Electrical and Computer Engineering, University of Ottawa, Ottawa, Ontario, Canada
        {\tt\small zeping.zhang@uottawa.com, laganier@uottawa.ca}}%
}

\begin{document}

\maketitle
\thispagestyle{empty}
\pagestyle{empty}


\begin{abstract}
LiDAR perception is severely limited by the distance-dependent sparsity of distant objects. While diffusion models can recover dense geometry, they suffer from prohibitive latency and physical hallucinations manifesting as ghost points. We propose Scanline-Consistent Range-Aware Diffusion, a framework that treats densification as probabilistic refinement rather than generation. By leveraging Partial Diffusion (SDEdit) on a coarse prior, we achieve high-fidelity results in just 156\,ms. Our novel Ray-Consistency loss and Negative Ray Augmentation enforce sensor physics to suppress artifacts. Our method achieves state-of-the-art results on KITTI-360 and nuScenes, directly boosting off-the-shelf 3D detectors without retraining. Code will be made available.
\end{abstract}

\section{INTRODUCTION}
\label{sec:intro}

\begin{figure}[]
\centering
\includegraphics[width=0.95\columnwidth]{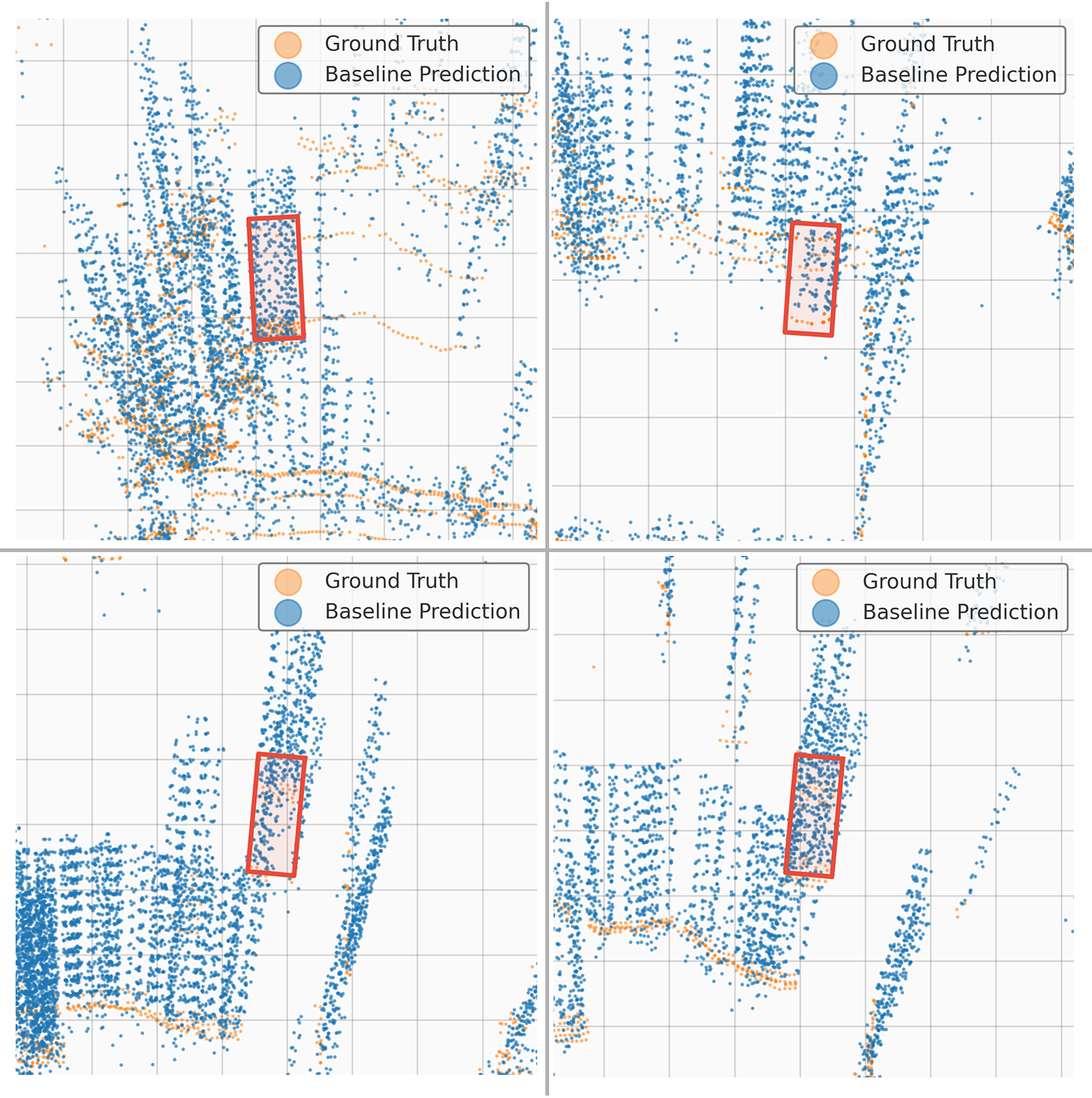}
\caption{\textbf{Bleeding Artifacts in LiDAR Densification.} An example of a densified result generated by the baseline method LiDPM. Note that without physical constraints, existing methods often generate severe ghost artifacts extending behind objects into free space.}
\label{fig:longtail}
\end{figure}

LiDAR-based 3D perception is fundamentally limited by sensor resolution. The quadratic decay of point density with distance results in severe sparsity for distant objects. To address this, generative densification methods have been proposed, but they often hallucinate geometry, creating bleeding artifacts or 'ghost trails' (see Fig.~\ref{fig:longtail}) behind objects where the sensor beam should pass through unimpeded. These hallucinated artifacts severely degrade the performance of downstream tasks like 3D object detection~\cite{yin2021center, chen2023voxelnext}.

To address this, LiDAR densification aims to recover dense geometry from sparse inputs. While deterministic methods (e.g., depth completion~\cite{ku2018defense}) are efficient, they often produce over-smoothed results that fail to capture high-frequency geometric details. Recently, Denoising Diffusion Probabilistic Models (DDPMs)~\cite{ho2020denoising} have achieved state-of-the-art fidelity in 3D generation~\cite{zyrianov2022learning, nunes2024lidiff}. However, applying standard diffusion to real-world perception pipelines faces two unresolved conflicts:

\textbf{1. The Efficiency Gap:} Standard diffusion requires iterative denoising from pure Gaussian noise (often $1000+$ steps), resulting in high latency (few seconds per frame) that is prohibitive for real-time systems~\cite{nunes2024lidiff}.

\textbf{2. The Physics Gap:} LiDAR data is governed by ray-casting physics. Existing generative models often treat points as generic coordinates, leading to "hallucinations" (ghost points) in known free space~\cite{zyrianov2022learning,pidm2025}. These artifacts, while visually plausible, can trigger phantom braking in autonomous systems.

In this work, we propose Scanline-Consistent Range-Aware Diffusion, a framework that bridges the gap between generative fidelity and physical constraints. We argue that densification should be treated as probabilistic refinement rather than generation from scratch. By initializing with a coarse structural prior and employing Partial Diffusion (SDEdit)~\cite{meng2021sdedit}, we reduce the search space significantly.

Our contributions are:
\begin{itemize}
    \item \textbf{Efficiency:} We propose a "Densify Before You Detect" front-end using Partial Diffusion, achieving high-fidelity densification in just 156ms.
    \item \textbf{Physics-Awareness:} We introduce Negative Ray Augmentation and Isotropic Probabilistic Consistency, forcing the model to respect sensor line-of-sight and eliminate ghost points.
    \item \textbf{Performance:} We achieve SOTA results on KITTI-360 and nuScenes. Crucially, our densified clouds can improve off-the-shelf 3D detection (such as VoxelNeXt~\cite{chen2023voxelnext}, CenterPoint~\cite{yin2021center}) without retraining.
\end{itemize}

\begin{figure*}[t]
\centering
\includegraphics[width=\textwidth, trim=0 15mm 0 0, clip]{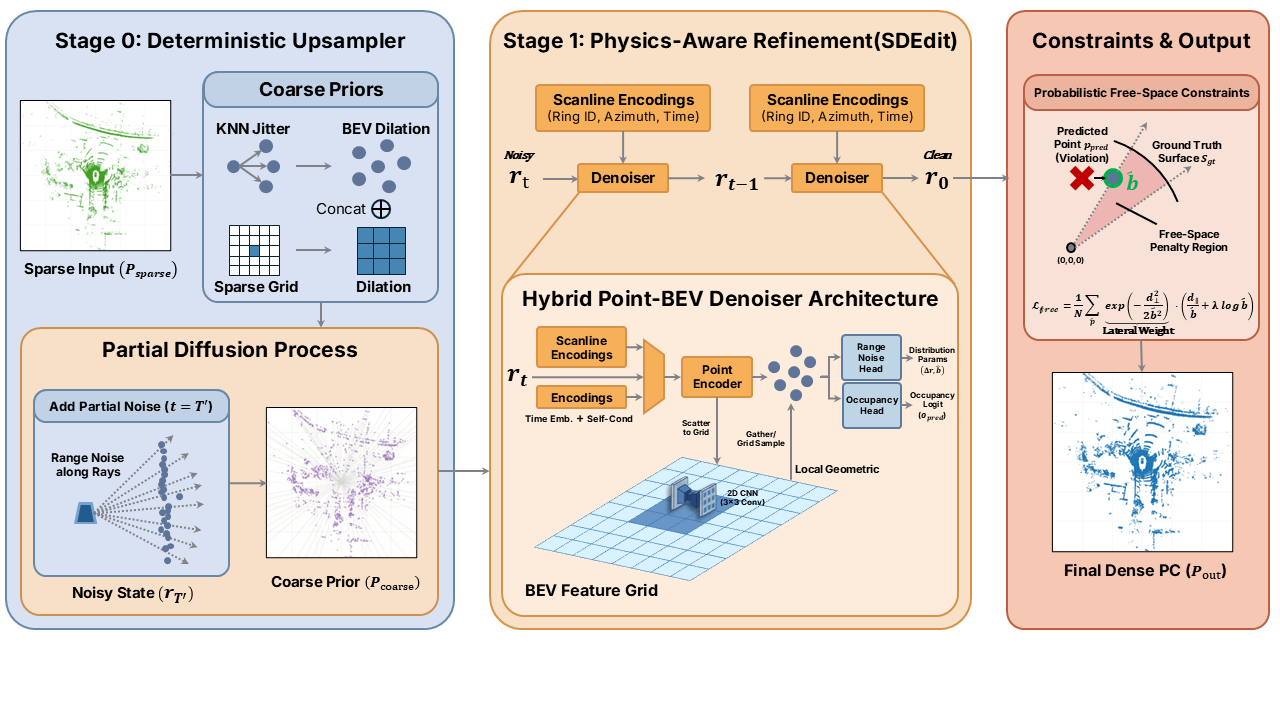}
\vspace{-6mm}
\caption{
\textbf{The proposed Scanline-Consistent Range-Aware Diffusion framework.}
(Left) Stage-0 constructs a coarse structural prior.
(Middle) Stage-1 performs Partial Diffusion. The Probabilistic Range Head predicts both range noise and an uncertainty scale $\hat{b}$.
(Right) Physics-Aware Constraints. We employ an \textbf{Isotropic Probabilistic Consistency} loss.
The penalty is weighted by a spatial Gaussian kernel (Lateral Weight) scaled by $\hat{b}$, allowing the uncertainty to dynamically determine the valid geometric tolerance for both lateral misalignment ($d_{\perp}$) and radial occlusion ($d_{\parallel}$).
}
\label{fig:main_arch}
\vspace{-4mm}
\end{figure*}

\section{RELATED WORK}
\label{sec:related}

\subsection{Deterministic Completion}
Early methods treated densification as a regression problem. Non-learning approaches like IP-Basic~\cite{ku2018defense} use morphological operations for fast upsampling but lack semantic coherence. Deep learning approaches, including sparse-to-dense networks, regress missing depth values. While computationally inexpensive, these methods minimize mean-squared error, thus inherently favoring blurry, mean-seeking predictions that degrade object boundaries.

\subsection{Generative LiDAR Modeling}
To recover sharp details, recent works leverage generative models. PVD~\cite{zhou2021pvd} introduced point-voxel diffusion for object completion. Scaling to scenes, LiDARGen~\cite{zyrianov2022learning} and LiDiff~\cite{nunes2024lidiff} utilize stochastic denoising to hallucinate realistic structures. More recently, the field has expanded toward foundational world models using latent flow matching~\cite{liutianran1} and vision-guided feature diffusion~\cite{liutianran2}. However, these methods often suffer from high inference latency and lack explicit free-space constraints, leading to "ghost points" in open regions. In contrast to recent works such as DiffRefine~\cite{shin2025diffrefine} that focuses on proposal-level densification, we target full-scene consistency.
While concurrent residual-based approaches like ResGen\cite{resgen2026}  attempt to improve stability, by formulating refinement using unconstrained 3D residual vectors generation. This formulation still permits points to drift into physically invalid free space. In contrast, our method employs a physics-constrained Range-Only Diffusion that strictly updates scalar depths along fixed sensor rays, enforcing geometric validity via our Ray Consistency Loss. 

\subsection{Our Approach}
We diverge from previous works by formulating densification as a \textit{conditional refinement} task. Instead of generating from noise, we adopt the SDEdit paradigm~\cite{meng2021sdedit}, effectively "warm-starting" the diffusion process with a deterministic prior. This allows the model to focus on high-frequency geometric recovery (improving efficiency) while our novel ray-consistency losses ensure that the output adheres to sensor physics (improving safety).

\section{Methodology}
\label{sec:method}

\subsection{Problem Formulation: Range-Manifold Densification}
\label{subsec:problem}
Given a sparse LiDAR sweep $\mathcal{P}_{in} = \{p_i\}_{i=1}^{N_{in}}$, our goal is to generate a dense, physically valid point cloud $\mathcal{P}_{out}$ that recovers the underlying geometry.
Unlike previous works that treat point cloud completion as a generic 3D generation problem $(x,y,z) \sim p_\theta(x)$, we leverage the projective nature of LiDAR sensors.
We decompose any point $p \in \mathbb{R}^3$ into a ray direction $\mathbf{d} \in S^2$ and a scalar range $r \in \mathbb{R}^+$.

Instead of a deterministic regression, We parameterize the predictive distribution along each ray as a Laplacian $p(r|\mathbf{d}) = \frac{1}{2\hat{b}} \exp(-\frac{|r - \mu|}{\hat{b}})$. Here, $r$ represents the estimated depth (geometry), while $b$ represents the aleatoric uncertainty (confidence).
Crucially, as visualized in Fig.~\ref{fig:main_arch} (Right), we interpret $\hat{b}$ not merely as 1D range noise, but as an isotropic uncertainty scale in 3D Euclidean space.
This allows the model to learn a flexible spatial tolerance: points with high uncertainty are permitted to exhibit larger lateral and radial fluctuations, while high-confidence predictions must strictly adhere to the sensor's line-of-sight geometry.

\subsection{Architecture Overview}
Our framework, illustrated in Figure~\ref{fig:main_arch}, adopts a coarse-to-fine strategy comprising two stages:
\begin{itemize}
    \item \textbf{Stage-0 (Deterministic Structural Prior):} Efficiently aggregates a coarse set of candidate rays $\mathcal{P}_{coarse}$ to define the query directions $\mathbf{d}$. 
    \item \textbf{Stage-1 (Physics-Aware Range Refinement):} Utilizes a conditional diffusion model to refine the range values $r$ along these directions.
\end{itemize}

A key aspect of our method is that, instead of generating from pure Gaussian noise, we employ a \textit{Partial Diffusion} (SDEdit, Stochastic Differential Editing) process initialized from the coarse prior, ensuring high fidelity and low latency.

\begin{figure}[t]
  \centering
  \includegraphics[width=1.0\columnwidth]{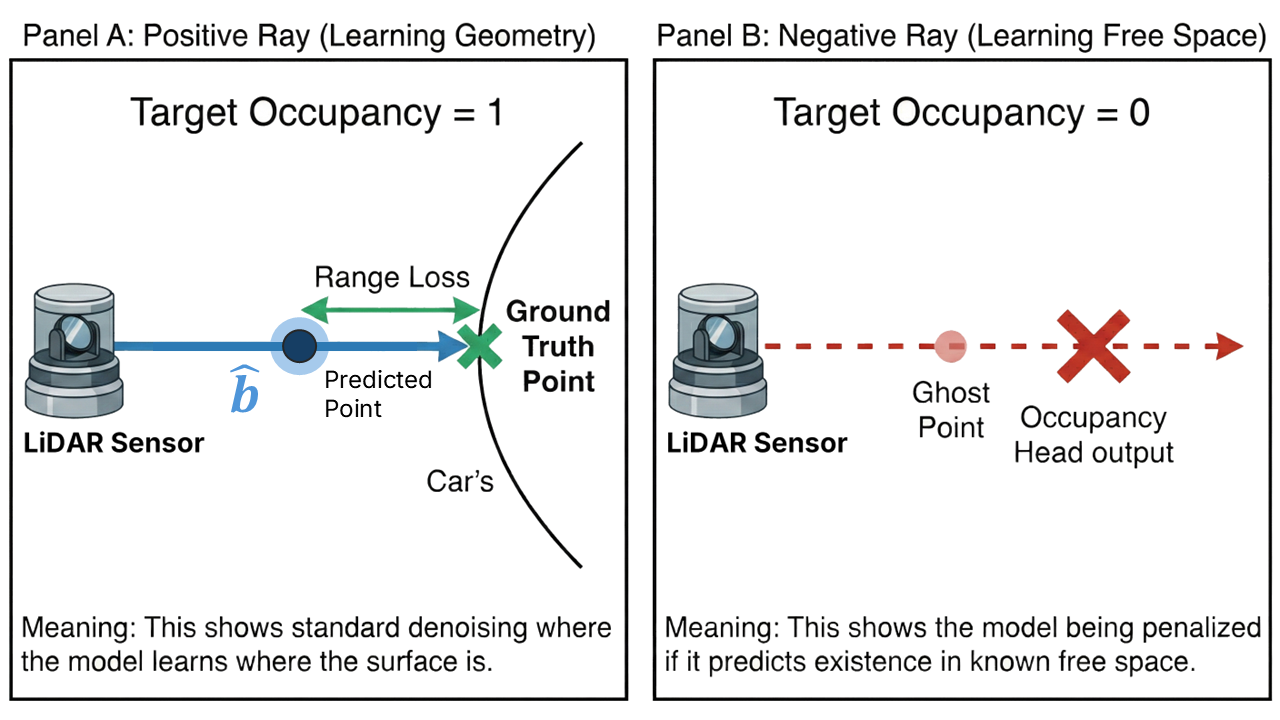} 
  \vspace{-4mm}
\caption{
\textbf{Negative Ray Augmentation Strategy.}
(Panel A) On positive rays, the model learns a probabilistic range distribution. The uncertainty $\hat{b}$ (visualized as the green confidence region) represents the isotropic spatial tolerance learned by the network.
(Panel B) On negative rays sampled from known free space, we enforce a zero-occupancy constraint.
This explicitly teaches the Occupancy Head to suppress ghost points in free space where the sensor beam should pass through unimpeded.
}
\label{fig:neg_ray}
  \vspace{-4mm}
\end{figure}

\subsection{Stage-0: Deterministic Structural Prior}
\label{subsec:stage0}
The first stage generates a structural skeleton $\mathcal{P}_{coarse}$ that serves as both the geometric condition and the starting point for diffusion.
To handle the sparsity of single-sweep LiDAR, we employ two complementary heuristics:
\begin{enumerate}
    \item KNN Jittering: To densify local surfaces, we sample $M$ neighbors for each input point $p_i$ via a Gaussian kernel $\mathcal{N}(p_i, \sigma_{jit}^2)$, capturing local surface variations.
    \item BEV Morphological Expansion: To recover missing structures (e.g., occluded vehicle parts), we project the raw sparse input $\mathcal{P}_{in}$ onto a binary bird-eye-view(BEV) grid, apply morphological dilation to close holes, and back-project active cells to 3D using local height statistics from $\mathcal{P}_{in}$.
\end{enumerate}
The union of these sets forms $\mathcal{P}_{coarse}$. Note that $\mathcal{P}_{coarse}$ is noisy and artifact-prone; its primary role is to provide a comprehensive set of candidate ray directions $\mathbf{d}_{coarse} = \vec{p} / \|\vec{p}\|$ for the subsequent refinement stage.

\subsection{Stage-1: Hybrid Point-BEV Range Diffusion}
\label{subsec:stage1}

\begin{figure*}[!t]
\centering
\includegraphics[width=\textwidth]{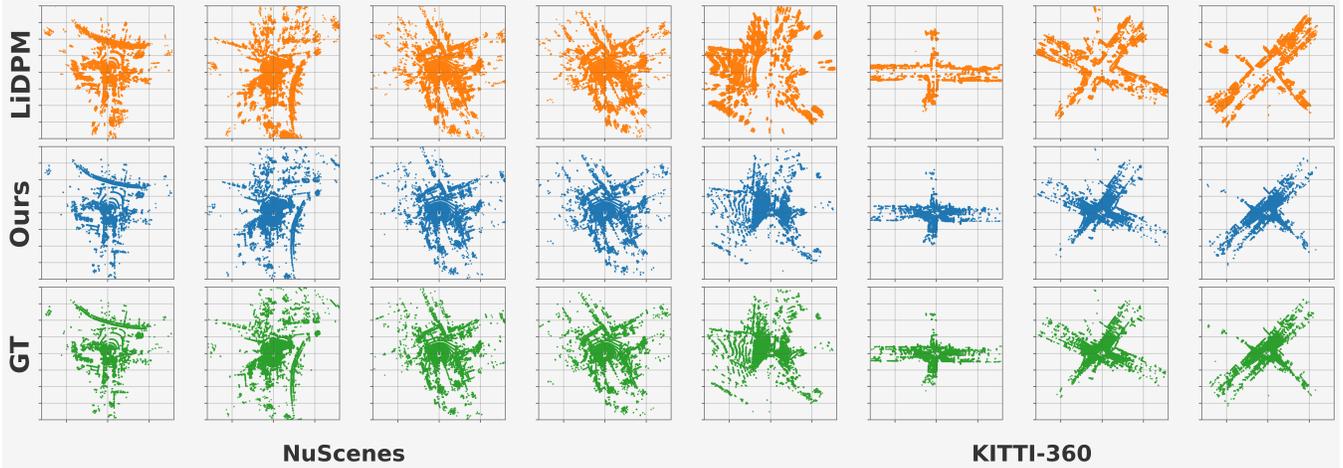}
\caption{\textbf{Qualitative Comparison of Densification Results.}
LiDPM preserves local detail but introduces scattered outlier points near object edges and occasional ghost points in free space. Our method generates sharp structures with clean free space in the demonstration.
}
\label{fig:qualitative}
\end{figure*}

\paragraph{Partial Diffusion via SDEdit.}
Standard diffusion models denoise from a standard Gaussian distribution $\mathcal{N}(0, I)$, requiring hundreds of steps to form structure.
Since we already start from a reasonable initialization $\mathcal{P}_{coarse}$, we adopt the SDEdit formulation.
We define the range state $r_0 = \|p_{coarse}\|$. We first diffuse $r_0$ to an intermediate timestep $T' < T$ by adding Gaussian noise:
\begin{equation}
    r_{T'} = \sqrt{\bar{\alpha}_{T'}} r_0 + \sqrt{1 - \bar{\alpha}_{T'}} \epsilon, \quad \epsilon \sim \mathcal{N}(0, I)
\end{equation},
where $\bar{\alpha}_{t'}$ is the cumulative noise schedule parameter at timestep $t'$.
The neural network then performs the reverse denoising process from $t=T'$ to $0$.
This \textit{Range-Only Partial Diffusion} focuses the model's capacity on refining geometry rather than generating global structure from scratch.

\paragraph{Hybrid Point-BEV Architecture.}
To effectively process LiDAR data which is sparse in 3D but structured in BEV, we propose a Scatter-Conv-Gather architecture.
The point encoder processes individual points augmented with scanline encodings (Ring ID, Azimuth, Time).
Simultaneously, a BEV branch facilitates long-range interaction:
(1) Scatter: Point features are projected onto a high-resolution BEV feature grid using max-pooling within grid cells.
(2) Conv: A lightweight 2D CNN processes the grid to propagate local context and complete shapes.
(3) Gather: The enhanced grid features are bilinearly sampled back to the 3D points.
This hybrid design captures fine-grained depth-wise details via the point branch and holistic scene structure via the BEV branch.

\paragraph{Dual-Head Prediction.}
The network terminates with two specialized heads:
\begin{itemize}
    \item \textbf{Probabilistic Range Head:} Predicts both the denoising component $\hat{\epsilon} = \epsilon_\theta(r_t, t, \mathcal{C})$ and the aleatoric uncertainty parameter $\hat{b}$.
    \item \textbf{Occupancy Head:} Predicts a logit $o \in \mathbb{R}$ indicating whether the ray hits a valid surface or traverses free space. This is critical for filtering out ghost points generated by the Stage-0 dilation.
\end{itemize}

\subsection{Physics-Aware Training Objectives}
\label{subsec:losses}

Our training strategy is strictly grounded in LiDAR physics. We optimize a composite objective $\mathcal{L} = \mathcal{L}_{diff} + \lambda_{occ}\mathcal{L}_{occ} + \lambda_{free}\mathcal{L}_{free}$.

\paragraph{Range Denoising Loss.}
Standard simple diffusion loss applied only to the scalar range channel:
\begin{equation}
    \mathcal{L}_{diff} = \mathbb{E}_{t, \epsilon} \left[ \| \epsilon - \epsilon_\theta(r_t, t, \mathcal{C}) \|^2 \right]
\end{equation},
where $r_t$ represents the latent range state at timestep $t$, initialized from the coarse prior $r_0$, and $\mathcal{C} = \{ \mathcal{P}_{coarse}, \mathbf{E}_{scan} \}$ represents the conditioning context, comprising the coarse prior and scanline encodings.

\paragraph{Occupancy and Negative Ray Augmentation.}
A common failure mode in densification is "hallucination" in free space. To mitigate this, we introduce a \textbf{Negative Ray Augmentation} strategy during training.
For each batch, we replace a fraction of valid rays with randomly sampled negative rays directing into known free space.
We enforce a Binary Cross Entropy (BCE) loss on the Occupancy Head:
\begin{equation}
    \mathcal{L}_{occ} = \text{BCE}( \sigma(o_{pred}), \mathbb{1}_{valid} )
\end{equation}
where $\mathbb{1}_{valid}$ is 1 for real points and 0 for negative rays. This explicitly teaches the model to distinguish between valid geometry and empty space.

\paragraph{Isotropic Probabilistic Consistency.}
Ideally, the uncertainty scale $\hat{b}$ should represent a 3D geometric confidence region.
Instead of applying hard angular thresholds, we formulate a fully probabilistic constraint based on spatial soft-association.
We interpret the predicted point as a probabilistic volume centered at $\hat{p}$ with scale $\hat{b}$.
For a neighbor ray $R_{in}^*$, we calculate two geometric metrics shown in Fig.~\ref{fig:main_arch}: the lateral distance $d_{\perp}$ (point-to-line deviation) and the radial occlusion depth $d_{\parallel} = \text{ReLU}(r_{in}^* - \hat{r}_{proj})$.
The free-space loss is weighted by the spatial alignment probability:
\begin{equation}
    \mathcal{L}_{free} = \frac{1}{N} \sum_{\hat{p}} \underbrace{\exp\left(-\frac{d_{\perp}^2}{2\hat{b}^2}\right)}_{\text{Lateral Weight}} \cdot \left( \frac{d_{\parallel}}{\hat{b}} + \lambda \log \hat{b} \right)
\end{equation}
Here, the exponential term acts as a learned attention mechanism:
(1) If the model is uncertain ($\hat{b}$ is large), it tolerates larger lateral deviations $d_{\perp}$ without enforcing the occlusion penalty.
(2) If the model is confident ($\hat{b}$ is small), the spatial weight vanishes unless the point is perfectly aligned with the ray.
This effectively treats $\hat{b}$ as the isotropic geometric scale, unifying both radial and lateral consistency.

\paragraph{Curriculum Density Schedule.}
To stabilize training, we employ a curriculum schedule $\rho(e)$ that linearly increases the target point density ratio from $\times 2$ to $\times 8$ over the training epochs. This prevents the model from collapsing when attempting to learn high-frequency details early in training.

\begin{table*}[t]
\centering
\caption{\textbf{Quantitative Comparison on KITTI-360 and nuScenes Validation Sets.}
$\downarrow$ indicates lower is better.
Best results are in \textbf{bold} and second-best are \underline{underlined}.
PVD and PCDM are capped at 8K points due to attention memory limits.
}
\label{tab:main_results}
\resizebox{\textwidth}{!}{%
\begin{tabular}{l|c||cccc|cccc}
\hline
& & \multicolumn{4}{c|}{\textbf{KITTI-360} (64-beam, ${\sim}$120K $\to$ ${\sim}$250K)} & \multicolumn{4}{c}{\textbf{nuScenes} (32-beam, ${\sim}$34.5K $\to$ ${\sim}$150K)} \\
\textbf{Method} & \textbf{Venue}
& \textbf{CD\,(m)} $\downarrow$ & \textbf{EMD} $\downarrow$ & \textbf{FSVR\,(\%)} $\downarrow$ & \textbf{REAP\,(\%)} $\downarrow$
& \textbf{CD\,(m)} $\downarrow$ & \textbf{EMD} $\downarrow$ & \textbf{FSVR\,(\%)} $\downarrow$ & \textbf{REAP\,(\%)} $\downarrow$ \\
\hline
\hline
\textit{Non-Generative} & & & & & & & & & \\
IP-Basic~\cite{ku2018defense} & IROS'18 
& 0.2614 & 0.3427 & 4.87 & 10.42 
& 0.2973 & 0.3846 & 5.21 & 11.63 \\
\hline
\textit{Object-level Diffusion} & & & & & & & & & \\
PVD~\cite{zhou2021pvd} & ICCV'21 
& 0.3782 & 0.4965 & 11.53 & 17.28 
& 0.4136 & 0.5374 & 12.74 & 18.95 \\
PCDM~\cite{pcdm2025} & NPL'25  
& 0.2735 & 0.3581 & 7.64  & 12.35 
& 0.3024 & 0.3918 & 8.26  & 13.41 \\
\hline
\textit{LiDAR Scene Generation} & & & & & & & & & \\
LiDARGen~\cite{zyrianov2022learning} & ECCV'22 
& 0.2948 & 0.3852 & 8.42 & 13.18 
& 0.3287 & 0.4243 & 9.18 & 14.27 \\
R2DM~\cite{nakashima2024r2dm} & ICRA'24 
& 0.2416 & 0.3169 & 5.37 & 7.95  
& 0.2745 & 0.3521 & 5.94 & 8.67 \\
LiDM~\cite{ran2024lidm} & CVPR'24 
& 0.1937 & 0.2584 & 4.92 & 7.46  
& 0.2193 & 0.2907 & 5.46 & 8.13 \\
\hline
\textit{Scene Completion / General} & & & & & & & & & \\
LiDPM~\cite{zyrianov2024lidpm} & IV'25  
& \underline{0.1563} & \underline{0.2158} & \underline{3.84} & \underline{5.91}  
& \underline{0.1785} & \underline{0.2436} & \underline{4.21} & \underline{6.58} \\
LiDiff~\cite{nunes2024lidiff} & CVPR'24 
& 0.2189 & 0.2876 & 5.71 & 9.23  
& 0.2478 & 0.3194 & 6.37 & 10.12 \\
\hline
\textbf{Ours} & \textbf{---} 
& \textbf{0.1512} & \textbf{0.1734} & \textbf{2.77} & \textbf{5.27} 
& \textbf{0.1671} & \textbf{0.2293} & \textbf{3.96} & \textbf{6.21} \\
\hline
\end{tabular}%
}
\vspace{-2mm}
\end{table*}

\section{EXPERIMENTS}
\label{sec:experiments}

\subsection{Experimental Setup}

\textbf{Datasets.}
We use two autonomous-driving benchmarks: \textbf{KITTI-360}~\cite{liao2023kitti360} (Velodyne HDL-64E, 64 beams, ${\sim}$120K pts/sweep; sequence 03 for validation~\cite{zyrianov2022learning}; dense GT: 10-sweep accumulation, ${\sim}$250K pts, 2.1$\times$ ratio) and \textbf{nuScenes}~\cite{caesar2020nuscenes} (HDL-32E, 32 beams, ${\sim}$34.5K pts/sweep; official 700/150/150 split; dense GT: 5-sweep aggregation~\cite{yin2021center}, ${\sim}$150K pts, 4.3$\times$ ratio).
Table~\ref{tab:data_stats} gives point-count statistics.

\begin{table}[t]
\centering
\caption{\textbf{Dataset Statistics.}
Avg.\ point counts for sparse input (single sweep), dense ground truth (multi-sweep accumulation), and our model's typical output.
}
\label{tab:data_stats}
\resizebox{\columnwidth}{!}{%
\begin{tabular}{l|ccc|c}
\hline
\textbf{Dataset} & \textbf{Sparse Input} & \textbf{Generated (Ours)} & \textbf{Dense GT} & \textbf{Ratio} \\
\hline
KITTI-360 (64-beam) & ${\sim}$120K & ${\sim}$240K & ${\sim}$250K & 2.1$\times$ \\
nuScenes (32-beam)  & ${\sim}$34.5K & ${\sim}$142K & ${\sim}$150K & 4.3$\times$ \\
\hline
\end{tabular}%
}
\end{table}

\textbf{Metrics.}
We report chamfer distance (CD, in meters) and earth mover's distance (EMD) for geometry, Free-Space Violation Ratio (FSVR) for free-space safety (ghost points), and Relative Error of Average number of Points (REAP) \cite{resgen2026} (\%) for density consistency. To calculate FSVR and account for the mathematical impossibility of exact point-to-ray intersection in continuous space, a generated point is classified as a free-space violation if its lateral distance to a ground-truth ray is strictly within a tolerance threshold of 0.1 m, and its radial depth is shorter than the ground-truth return. This 0.1 m threshold is empirically chosen to accommodate the physical beam divergence of typical LiDAR sensors and the slight registration noise inherent in multi-sweep ground-truth accumulation.

\textbf{Baselines.}
We compare against PVD~\cite{zhou2021pvd}, PCDM~\cite{pcdm2025}, LiDARGen~\cite{zyrianov2022learning}, R2DM~\cite{nakashima2024r2dm}, LiDM~\cite{ran2024lidm}, LiDPM~\cite{zyrianov2024lidpm}, LiDiff~\cite{nunes2024lidiff}, and IP-Basic~\cite{ku2018defense}. Some implementation details are given in Table \ref{tab:implementation}.

\begin{table}[ht]
\centering
\caption{Implementation details of the proposed method.}
\label{tab:implementation}
\begin{tabular}{llc}
\toprule
\textbf{Category} & \textbf{Parameter} & \textbf{Value} \\
\midrule
\multirow{3}{*}{Architecture} & Hidden size $H$ & 256 \\
                              & Layers & 6 \\
                              & BEV Resolution & $64 \times 64$ \\
\midrule
\multirow{5}{*}{Training}     & Total Epochs & 200 \\
                              & Batch Size & 8 \\
                              & Optimizer & AdamW ($lr=10^{-4}$) \\
                              & Schedule & Cosine ($T=1000$) \\
                              & Self-conditioning & 0.5 \\
\midrule
\multirow{2}{*}{Data/Stage}   & Curriculum Density & $2\times \to 8\times$ \\
                              & Stage-0 ($k$, Jitter) & 8, 0.10\,m \\
\midrule
\multirow{2}{*}{Inference}    & DDIM Steps & 50 \\
                              & SDEdit $\alpha$ & 0.25 \\
\bottomrule
\end{tabular}
\end{table}

\begin{table*}[]
\centering
\small
\caption{\textbf{Computational Efficiency.}
We evaluate all methods with batch size 1.
$^\dagger$PVD encountered OOM when processing 150K points.
}
\label{tab:efficiency}
\begin{tabular}{l|cccccccc}
\hline
\textbf{Method} 
& \textbf{Ours} 
& \textbf{LiDiff}$^\ddagger$ 
& \textbf{LiDPM} 
& \textbf{R2DM} 
& \textbf{LiDARGen} 
& \textbf{PVD}$^\dagger$ 
& \textbf{LiDM} 
& \textbf{IP-Basic} \\
\hline
\#Params 
& 8.6M 
& 38.2M 
& 50.7M 
& 45.5M 
& 80.8M 
& 13.6M 
& 75.6M 
& CPU \\
Infer.\,(ms) $\downarrow$ 
& \textbf{156} 
& 2{,}340 
& 1{,}559 
& 434 
& 823 
& OOM 
& 20 
& \textbf{13} \\
\hline
\end{tabular}%
\end{table*}

\begin{table}[t]
\centering
\caption{\textbf{Ablation Study on KITTI-360 and nuScenes.}
Impact of the Ray Consistency Loss, Scanline Encoding, Coarse Initialization, and BEV Branch.
}
\label{tab:ablation}
\resizebox{\columnwidth}{!}{%
\begin{tabular}{l|cc|cc|c}
\hline
& \multicolumn{2}{c|}{\textbf{KITTI-360}} & \multicolumn{2}{c|}{\textbf{nuScenes}} & \\
\textbf{Configuration} & \textbf{CD} $\downarrow$ & \textbf{FSVR} $\downarrow$ & \textbf{CD} $\downarrow$ & \textbf{FSVR} $\downarrow$ & \textbf{Infer.\,(ms)} \\
\hline
\textbf{Full Model (Ours)} & \textbf{0.1512} & \textbf{2.77} & \textbf{0.1671} & \textbf{3.96} & \textbf{156} \\
\hline
w/o $\mathcal{L}_{\text{free}}$ (Ray Loss) & 0.179 & 9.0 & 0.196 & 12.4 & 156 \\
w/o Scanline Encoding        & 0.169 & 4.2 & 0.188 & 5.9 & 149 \\
w/o Stage-0 (Pure Noise Init) & 0.211 & 4.9 & 0.231 & 7.0 & 314 \\
w/o BEV Branch (Point-only)  & 0.193 & 4.4 & 0.209 & 6.1 & 122 \\
\hline
\end{tabular}%
}
\end{table}

\subsection{Quantitative Results}
\label{sec:quantitative}

Our method ranks first on all eight metric--dataset combinations. On KITTI-360, we achieve CD\,=\,0.1512\,m, which is 3.3\% lower than LiDPM (0.1563\,m).
On nuScenes (sparser 32-beam, 4.3$\times$ densification), CD\,=\,0.1671\,m, representing a 6.4\% reduction over LiDPM (0.1785\,m). These results demonstrate consistent gains across sensor types and sparsity regimes.

The largest gain is observed on FSVR. On KITTI-360, FSVR decreases from 3.84\% to 2.77\%, an absolute reduction of 1.07\% (27.9\% relative reduction).
On nuScenes, FSVR drops from 4.21\% to 3.96\%. Validating the Ray Consistency Loss $\mathcal{L}_\text{free}$. REAP (5.27\% / 6.21\%) is also best across methods.

\begin{table}[t]
\centering
\caption{\textbf{Downstream 3D Detection on nuScenes.}
Off-the-shelf SOTA detectors on raw sparse (1 sweep), LiDPM-densified, and our densified scans. All values are overall mAP and NDS.
}
\label{tab:downstream}
\resizebox{\columnwidth}{!}{%
\begin{tabular}{l|l|cc}
\hline
\textbf{Detector} & \textbf{Input Data} & \textbf{mAP} $\uparrow$ & \textbf{NDS} $\uparrow$ \\
\hline
\hline
\multirow{3}{*}{\textbf{VoxelNeXt}} 
& Raw Sparse (1 sweep)    & 61.2 & 68.4 \\
& Densified (LiDPM)      & 64.1 & 70.8 \\
& \textbf{Densified (Ours)} & \textbf{65.6} & \textbf{71.9} \\
\hline
\multirow{3}{*}{\textbf{CenterPoint}} 
& Raw Sparse (1 sweep)    & 56.3 & 64.5 \\
& Densified (LiDPM)      & 59.5 & 66.2 \\
& \textbf{Densified (Ours)} & \textbf{61.1} & \textbf{67.6} \\
\hline
\end{tabular}%
}
\end{table}

\subsection{Qualitative Analysis}
\label{sec:qualitative}

Due to limited space, we show only our method and the second-best baseline (LiDPM\cite{zyrianov2024lidpm}) on nuScenes and KITTI 360 (Figure~\ref{fig:qualitative}).
Our results show clean free space with fewer ghost points than LiDPM\cite{zyrianov2024lidpm}, sharp object boundaries (per-ray range denoising avoids cross-contamination), and consistent quality; LiDPM degrades more where input is sparser.

\subsection{Ablation Studies}
\label{sec:ablation}

We investigate the contribution of each key component.
All ablation experiments are conducted on the validation sets of both datasets using the same training configuration. The results are shown in TABLE \ref{tab:ablation}.

Removing $\mathcal{L}_\text{free}$ raises FSVR by ${\sim}$3$\times$ on both datasets and degrades CD by ${\sim}$18\%; removing Stage-0 degrades CD by ${\sim}$39\% and doubles inference time; removing BEV layers degrades CD by ${\sim}$25--27\% for only 34\,ms overhead; scanline encoding gives ${\sim}$10--12\% CD gain.

\subsection{Computational Efficiency}
\label{sec:efficiency}

Table~\ref{tab:efficiency} compares efficiency across all methods.
All timings are measured on a single NVIDIA RTX 4090 GPU with batch size 1, 50 DDIM steps, and 150K target points.

Ours achieves the lowest latency (156\,ms), smallest model (8.6M) thanks to 1D diffusion. While the non-learning baseline IP-Basic achieves lower latency through local interpolation, our generative approach justifies the computational cost by enabling the recovery of large-scale occlusions that are intractable for interpolation-based methods. (Table~\ref{tab:main_results}).

\subsection{Downstream Application: 3D Object Detection}
\label{sec:downstream}

We evaluate densification on nuScenes using two state-of-the-art LiDAR detectors: VoxelNeXt~\cite{chen2023voxelnext} (fully sparse, top-tier on nuScenes) and \textbf{CenterPoint}~\cite{yin2021center} (voxel-based anchor-free).
Detectors are applied off-the-shelf (no retraining on densified data).

Table~\ref{tab:downstream} reports overall mAP and NDS only (no per-class breakdown). The consistent gains across both detectors indicate that densification improves detection quality in aggregate. Our method yields higher mAP/NDS than LiDPM-densified input, which shows our point clouds quality matches the statistics detectors were trained on.

Densification consistently improves both detectors; gains are largest for distant/small objects (pedestrians, cones) where single-sweep data is sparse.
Our method yields higher mAP/NDS than LiDPM-densified input and avoids the false positives from LiDPM's free-space violations.
Off-the-shelf gains (no detector fine-tuning) show our point clouds match the statistics detectors were trained on.

\section{CONCLUSIONS}

\label{sec:conclusions}

In this paper, we presented a scanline-consistent, physics-aware diffusion framework for LiDAR densification. By formulating a two-stage coarse-to-fine pipeline and a range-only diffusion manifold, we successfully integrated sensor-specific scanline geometry and physical free-space constraints into a generative process. Our hybrid Point-BEV architecture and ray-consistency mechanism enable the recovery of sharp object structures while effectively suppressing hallucinations in empty space. Extensive evaluations on the KITTI-360 and nuScenes benchmarks demonstrate that our method achieves state-of-the-art performance in both geometric fidelity and free-space safety, with significantly lower latency than existing generative baselines. Furthermore, the consistent performance gains observed in off-the-shelf 3D detectors confirm the effectiveness of our approach as a robust and efficient front-end for autonomous driving perception stacks.






\bibliographystyle{plain}   
\bibliography{IEEEexample}   

\end{document}